\documentclass{article}
\usepackage[nonatbib,final]{neurips_2022}

\usepackage{amsmath,amsfonts,bm}

\def\eqref#1{equation~\ref{#1}}

\def\1{\bm{1}}

\def\vs{{\bm{s}}}

\DeclareMathAlphabet{\mathsfit}{\encodingdefault}{\sfdefault}{m}{sl}
\SetMathAlphabet{\mathsfit}{bold}{\encodingdefault}{\sfdefault}{bx}{n}

\def\gA{{\mathcal{A}}}

\def\gD{{\mathcal{D}}}
\def\gE{{\mathcal{E}}}

\def\gS{{\mathcal{S}}}
\def\gT{{\mathcal{T}}}
\def\gU{{\mathcal{U}}}

\usepackage{color,xcolor}
\usepackage{epsfig}
\usepackage{graphicx}

\usepackage{adjustbox}
\usepackage{array}
\usepackage{booktabs}
\usepackage{colortbl}
\usepackage{wrapfig}
\usepackage{hhline}
\usepackage{multirow}
\usepackage{subcaption} %
\usepackage{wrapfig}
\usepackage{floatflt}

\usepackage{amsmath,amsfonts,amssymb}
\usepackage{mathtools}  %
\usepackage{bm}
\usepackage{nicefrac}
\usepackage{microtype}

\usepackage{changepage}
\usepackage{extramarks}
\usepackage{fancyhdr}
\usepackage{lastpage}
\usepackage{setspace}
\usepackage{soul}
\usepackage{xspace}

\usepackage[pagebackref=true,breaklinks=true,colorlinks,citecolor=gray]{hyperref}

\usepackage{url}

\usepackage{enumerate}
\usepackage{todonotes} %
\usepackage{enumitem}  %

\usepackage{titlesec}

\usepackage{makecell}

\usepackage{pifont} %

\usepackage{algorithm,algpseudocode}

\usepackage{amsthm}
\usepackage{framed}
\definecolor{shadecolor}{gray}{0.95}
\newcolumntype{L}[1]{>{\raggedright\let\newline\\\arraybackslash\hspace{0pt}}m{#1}}
\newcolumntype{C}[1]{>{\centering\let\newline\\\arraybackslash\hspace{0pt}}m{#1}}
\newcolumntype{R}[1]{>{\raggedleft\let\newline\\\arraybackslash\hspace{0pt}}m{#1}}

\newcommand{\sect}[1]{Section~\ref{#1}}
\newcommand{\sectapp}[1]{Appendix~\ref{#1}}

\newcommand{\fig}[1]{Fig.~\ref{#1}}
\newcommand{\tbl}[1]{Table~\ref{#1}}

\newcommand{\ignore}[1]{}

\makeatletter
\DeclareRobustCommand\onedot{\futurelet\@let@token\@onedot}
\def\@onedot{\ifx\@let@token.\else.\null\fi\xspace}

\def\eg{e.g\onedot} 
\def\ie{i.e\onedot} 
 
\def\etc{etc\onedot}
\def\vs{\emph{vs}\onedot}

\makeatother

\definecolor{MyDarkBlue}{rgb}{0,0.08,1}
\definecolor{MyDarkGreen}{rgb}{0.02,0.6,0.02}
\definecolor{MyDarkRed}{rgb}{0.8,0.02,0.02}
\definecolor{MyDarkOrange}{rgb}{0.40,0.2,0.02}
\definecolor{MyPurple}{RGB}{111,0,255}
\definecolor{MyRed}{rgb}{1.0,0.0,0.0}
\definecolor{MyGold}{rgb}{0.75,0.6,0.12}
\definecolor{MyDarkgray}{rgb}{0.66, 0.66, 0.66}

\newcommand{\model}{PDSketch\xspace}

\newcommand{\newtext}[1]{{#1}}

\newcommand{\slot}{\textbf{??}\xspace}
\newcommand{\astar}{\textit{A$^*$}\xspace}

\newcommand{\xhdr}[1]{\noindent\textbf{#1}}

\newcommand{\mycellc}[1]{\begin{tabular}[t]{@{}c@{}l}#1\end{tabular}}

\usepackage[utf8]{inputenc} %
\usepackage[T1]{fontenc}    %
\usepackage{natbib}

\title{PDSketch: Integrated Planning Domain\\ Programming and Learning}

\author{%
Jiayuan Mao$^{1}$\\
\And
Tomás Lozano-Pérez$^{1}$
\And
Joshua B. Tenenbaum$^{1,2,3}$
\And
Leslie Pack Kaelbling$^{1}$
\AND
$^1$ MIT Computer Science \& Artificial Intelligence Laboratory\\
$^2$ MIT Department of Brain and Cognitive Sciences\\
$^3$ Center for Brains, Minds and Machines
}

\begin{document}
\maketitle

\footnotetext{Correspondence to: jiayuanm@mit.edu. Project page: https://pdsketch.csail.mit.edu.}

\vspace{-1em}
\begin{abstract}
\vspace{-0.5em}
This paper studies a model learning and online planning approach towards building flexible and general robots. Specifically, we investigate how to exploit the {\it locality} and {\it sparsity} structures in the underlying environmental transition model to improve model generalization, data-efficiency, and runtime-efficiency. We present a new domain definition language, named PDSketch. It allows users to flexibly define high-level structures in the transition models, such as object and feature dependencies, in a way similar to how programmers use TensorFlow or PyTorch to specify kernel sizes and hidden dimensions of a convolutional neural network. The details of the transition model will be filled in by trainable neural networks. Based on the defined structures and learned parameters, PDSketch automatically generates domain-independent planning heuristics without additional training. The derived heuristics accelerate the performance-time planning for novel goals.
\end{abstract}
\vspace{-0.5em}
\section{Introduction}
\vspace{-0.5em}

A long-standing goal in AI is to build robotic agents that are flexible and general, able to accomplish a diverse set of tasks in novel and complex environments. Such tasks generally require a robot to generate long-horizon plans for novel goals in novel situations, by reasoning about many objects and the ways in which their state-changes depend on one another. A promising solution strategy is to combine model learning with online planning: the agent forms an internal representation of the environment's states and dynamics by learning from external or actively-collected data, and then applies planning algorithms to generate actions, given a new situation and goal at performance time.

There are two primary desiderata for a system based on model-learning and planning. First, the learning process should be {\it data efficient}, especially because of the combinatorial complexity of possible configuration of in the real world. Second, the learned model should be {\em computationally efficient}, making online planning a feasible runtime execution strategy.

A critical strategy for learning models that generalize well from small amounts of data and that can be deployed efficiently at runtime is to introduce inductive biases.  In image processing, we leverage translation invariance and equivariance by using convolutions. In graph learning, we leverage permutation invariance by using graph neural networks. Two essential forms of structure we can leverage in dynamic models of the physical world are {\it locality} and {\it sparsity}. Consider a robot picking an object up off a table.  At the object level, the operation only changes the states of the object being picked up and the robot, and only depends on a few other nearby objects, such as the table (local).  At the object feature level, only the configuration of the robot and pose of the object are changed, but their colors and frictional properties are unaffected (sparse). 

\begin{figure}[!tp]
    \centering
    \small
    \includegraphics[width=\textwidth]{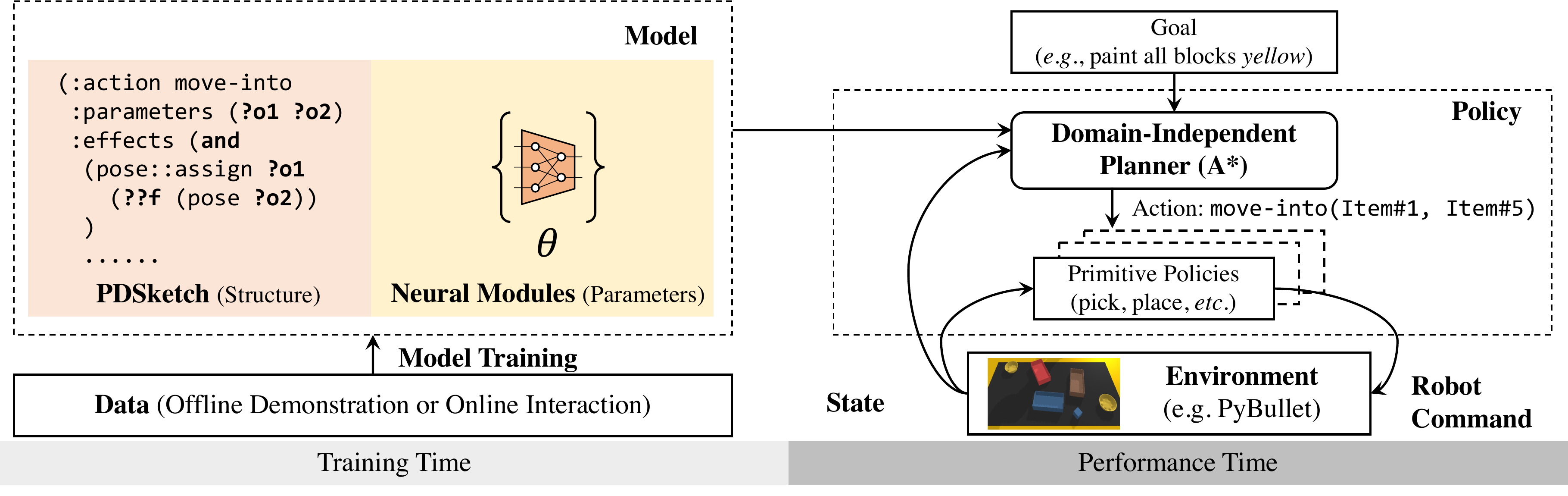}
    \vspace{-1.5em}
    \caption{The life cycle of a \model model. A \model model is composed of a model structure definition and a collection of trainable neural modules. The model parameters can be learned from data. During performance time, the model is used by a domain-independent planner to form a policy that directly interacts with the environment.}
    \label{fig:teaser}
    \vspace{-2em}
\end{figure} 

Classical hand-engineered approaches to robot task and motion planning have designed representations that expose and exploit locality through {\em lifting} (or {\em object-centrism}), which allows relational
\begin{wrapfigure}{r}{0.6\textwidth}
    \centering\small
    \includegraphics[width=0.59\textwidth]{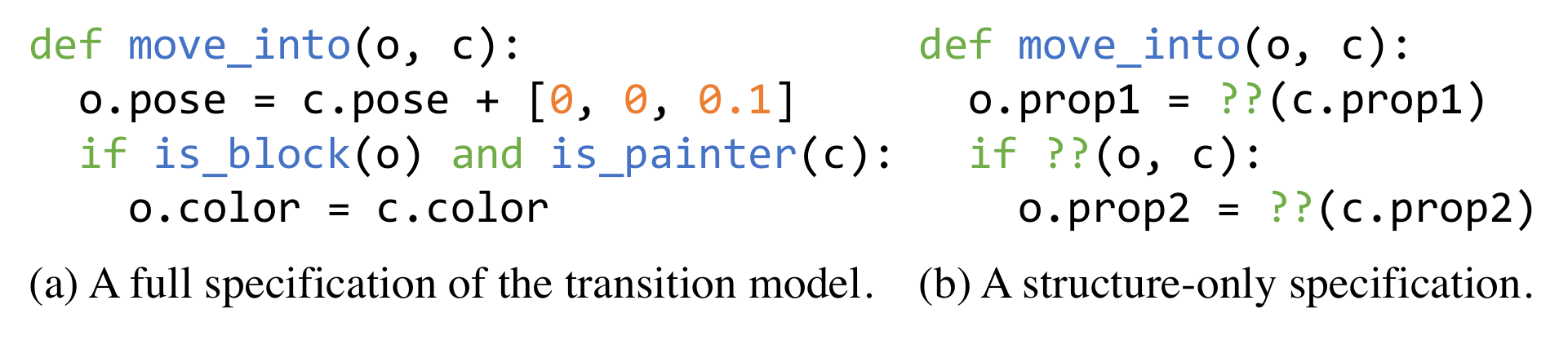}
    \caption{Defining both the transition model structure and implementation in Python (a) vs. defining only the structure while leaving details (the \slot functions) to be learned (b).}
    \label{fig:python}
\end{wrapfigure} %
descriptions of objects and abstraction over them, and through {\em factoring}, which represents different attributes of an object in a disentangled way~\citep{garrett2021integrated}. These representations are powerful and effective articulations of locality and sparsity, but they are traditionally laboriously hand-designed in a process that is very difficult to get correct, similar to writing a full state-transition function as in \fig{fig:python}a. This approach is not directly applicable to problems involving perception or environmental dynamics that are unknown or difficult to specify. In this paper, we present \model, a model-specification language that integrates human specification of structural sparsity priors and machine learning of continuous and symbolic aspects of the model. Just as human users may define the structure of a convolutional neural network in TensorFlow~\citep{tensorflow} or PyTorch~\citep{pytorch}, \model allows users to specify {\it high-level} structures of the transition model as in \fig{fig:python}b (analogous to setting the kernel sizes), and uses machine learning to fill in the details (analogous to learning the convolution kernels). 

\fig{fig:teaser} depicts the life-cycle of a PDSketch model, \model uses an object-centric, factored, symbolic language to flexibly describe structural inductive biases in planning domains (\ie, the model structure).  A \model model is associated with a collection of neural modules whose parameters can be learned from robot trajectory data that are either collected offline by experts or actively-collected by interacting with the environment. During performance time, a \model is paired with a domain-independent planner, such as \astar, and as a whole forms a goal-conditioned policy. %
The planner receives the environmental state and the trained \model model, makes plans in an abstract action space, and invokes primitive policies that actually generate robot 
joint commands.

Compared to unstructured models, such as a single multi-layer perceptron that models the complete state transition monolithically, the structures specified in \model substantially improve model generalization and data-efficiency in training.  In addition, they enable the computation of powerful {\it domain-independent planning heuristics}: these are estimates of the cost-to-go from each state to a state satisfying the goal specification, which can be obtained from the structured transition model {\it without any additional learning}. They can be leveraged by \astar to efficiently plan for {\it unseen} goals, specified in a first-order logic language.

We experimentally verify the efficiency and effectiveness of \model in two domains: BabyAI, an 2D grid-world environment that focuses on navigation, and Painting Factory, a simulated table-top robotic environment that paints and moves blocks. Our results suggest that 1) locality and sparsity structures, specified economically in a few lines of code, can significantly improve the data efficiency of model learning; 2) the model learning and planning paradigm enables strong generalization to unseen goal specifications. Finally, the domain-independent heuristics automatically induced from the structures dramatically improve performance-time efficiency, especially for novel goal specifications.
\vspace{-1.5em}
\section{PDSketch}
\vspace{-0.5em}

\begin{figure}[!tp]
    \centering\small
    \includegraphics[width=\textwidth]{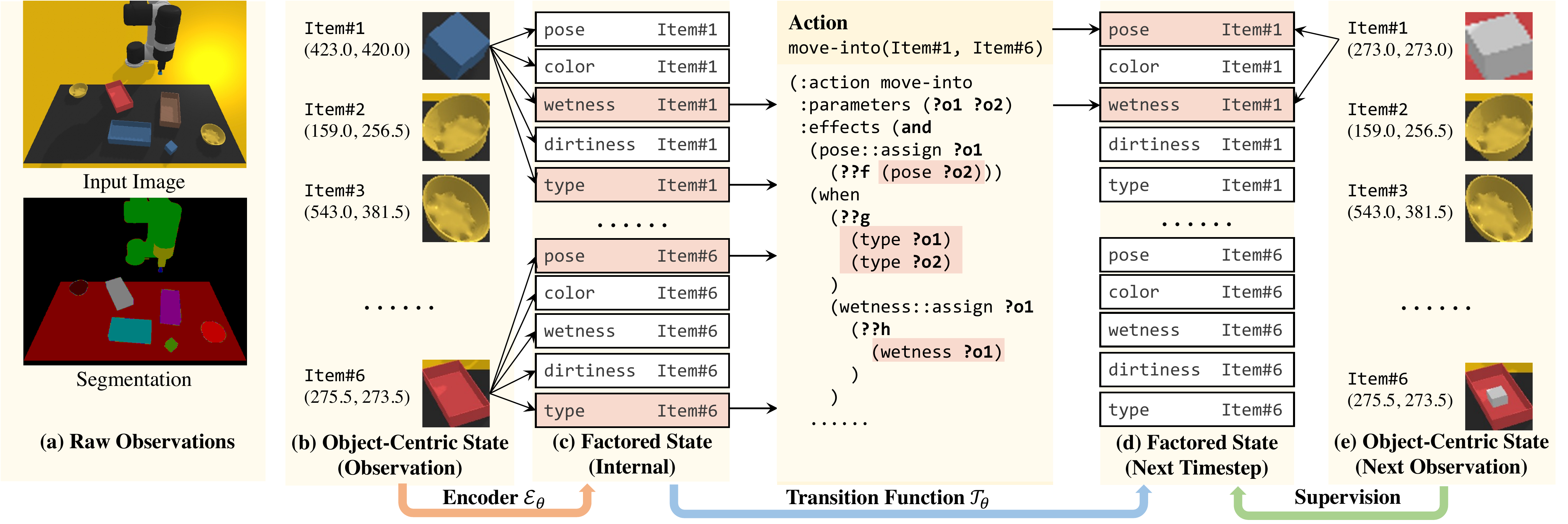}
    \caption{A factorized state representation and transition model for the robot painting domain. The raw observation (a) is first processed by an external perception module into an object-centric representation (b). This representation is further transformed into a fine-grained factorization (c). The transition function $\gT_\theta$ can be defined over this factored state representation: each action may only change a few factors of the state. Executing a specific action {\tt move-into(item\#1, item\#6)} produces the predicted factored state at the next timestep. During training time, we will be using the object-centric observation from the next timestep to supervise the learning of $\gT_\theta$.}
    \label{fig:factorized}
\end{figure} We focus on the problem of learning models for a robot that operates in a space $\gS$ of world states that plans to achieve goal conditions that are subsets of $\gS$. A planning problem is a tuple $\langle \gS, s_0, g, \gA, \gT \rangle$, where $s_0 \in \gS$ is the initial state, $g$ is a goal specification in first-order logic, $\gA$ is a set of actions that the agent can execute, and $\gT$ is a environmental transition model $\gT: \gS \times \gA \rightarrow \gS$. The task of planning is to output a sequence of actions $\bar{a} = \{a_i \in \gA\}$ in which the terminal state $s_T$ induced by applying $a_i$ sequentially following $\gT$ satisfies the goal specification $g$: $\textit{eval}(g, s_T) = 1$.  The function $\textit{eval}(g, s)$ determines whether state $s$ satisfies the goal condition $g$ by recursively evaluating the logical expression and using learned neural groundings of the primitive terms in the expression.

At execution time, the agent will observe $s_0$ and be given $g$ from human input, such as a first-order logic expression corresponding to ``all the apples are in a blue bowl.''  However, we do not assume that the agent knows, in advance, the groundings of $g$ (\ie the underlying $\textit{eval}(g, s)$ function) or the transition model $\gT$. Thus, we need to learn $g$ and $\gT$ from data, in the form of observed trajectories that achieve goal states of interest.  

Formally, we assume the training data given to the agent is a collection of tuples $\langle \overline{s}, \overline{a}, g, \overline{\textit{succ}}\rangle$, where $\overline{s} = \langle s_i\rangle$ is a sequence of world states, $\overline{a} = \langle a_i\rangle$ is the sequence of actions taken by the robot, $g$ is a goal specification, and   
$\overline{\textit{succ}} = \langle\textit{succ}_i\rangle$ is the ``task-success'' signal. Each $\textit{succ}_i \in \{0, 1\}$ indicates whether the goal $g$ is satisfied at state $s_i$: $\textit{succ}_i = \textit{eval}(g, s_i)$. 
The data sequences should be representative of the dynamics of the domain but need not be optimal goal-reaching trajectories.

It can be difficult to learn a transition model that is accurate over the long term on some types of state representations. For this reason, we generally assume an arbitrary latent space, $\Phi$, for planning. The learning problem, then, is to find three parametric functions, collectively parameterized by $\theta$: state encoder $\gE_\theta : \gS \rightarrow \Phi$, goal-evaluation function $\textit{eval}_{\theta} : \mathcal{G} \times \Phi \rightarrow \{0, 1\}$, and transition model $\gT_\theta: \Phi \times \gA \rightarrow \Phi$.  Although the domain might be mildly partially observable or stochastic, our goal will be to recover the most accurate possible deterministic model on the latent space.

\xhdr{Local, sparse structure.}
We need models that will generalize very broadly to scenarios with different numbers and types of objects in widely varying arrangements.
To achieve this, we exploit structure to enable compositional generalization:  throughout this work we will be committing to an object-centric representation for $s$, a logical language for goals $g$, and a sparse, local model of action effects.

We begin by factoring the environmental state $s$ into a set of object states. Each $s \in \gS$ is a tuple $(\gU_s, f_s)$, where $\gU_s$ is the set of objects in state $s$, denoted by arbitrary unique names (\eg, {\tt item\#1}, {\tt item\#2}). The object set $\gU_s$ is assumed to be constant within a single episode, but may differ in different episodes. The second component, $f_s$, is a dictionary mapping each object name to a fixed-dimensional object-state representation, such as a local image crop of the object and its position.  We can extend this representation to relations among objects, for example by adding  $g_s(x, y), x, y \in \gU$ as a mapping from each object pair to a vector representation. We assume the detection and tracking of objects through time is done by external perception modules (\eg, object detectors and trackers).

We carry the object-centric representation through to actions, goals, and the transition model.  Specifically, we define a {\it predicate} as a tuple $\langle \textit{name}, \textit{args}, \textit{grounding} \rangle$, where $\textit{args}$ is a list of $k$ arguments and $\textit{grounding}$ is a function from the latent representations of the objects corresponding to its arguments (in $\Psi^k$) into a scalar or vector value.  (This is a generalization of the typical use of the term "predicate," which is better suited for use in robotics domains in which many quantities we must reason about are continuous.)
For example, as illustrated in \fig{fig:factorized}c, the predicate {\tt wetness} takes a single argument as its input, and returns a (latent) vector representation of its wetness property; its grounding might be a neural network that maps from the visual appearance of the object to the latent wetness value.
Given a set of predicates, we define the language of possible goal specifications to be all first-order logic formulas over the subset of the predicates whose output type is Boolean.   To evaluate a goal specification in a state $(\gU, f)$, quantification is interpreted in the finite domain $\gU$ and $f_s$ provides an interpretation of object names into representations that can serve as input to the grounded predicates.

The transition model $\gT_\theta$ is specified in terms of a set object-parameterized action schema $\langle \textit{name}, \textit{args}, \textit{precond}, \textit{effect}, \pi\rangle$, where $\textit{name}$ is a symbol, $\textit{args}$ is a list of symbols, $\textit{precond}$ and $\textit{effect}$ are descriptions of the action's effects, described in section~\ref{sec:pdsketch}, and
$\pi$ is a parameterized primitive policy for carrying out the action in terms of raw perception and motor commands.  These local policies can be learned via demonstration or reinforcement learning in a phase prior to the model-learning phase, constructed using principles of control theory, or a combination of these methods.
The set of concrete actions $\gA$ available in a state $s$ is formed by instantiating the action scheme with objects in universe $\gU_s$.
We assume the transition dynamics of the domain (\ie, the \textit{effect} of each action schema) are well characterized in terms of the changes of properties and relations of objects and that the transition model is lifted in the sense that it can be applied to domain instances with different numbers and types of objects.
In addition, we assume the dynamics are local and sparse, in the sense that effects of any individual action depend on and change only a small number attributes and relations of a few objects, and that by default all other objects and attributes are unaffected.
Taking again action schema {\tt move-into} as an example, shown in \fig{fig:factorized}d, only the states of object \#1 and \#6 are relevant to this action (but not \#2, \#3, \etc), and furthermore, the action only changes the {\tt pose} and {\tt wetness} properties of {\tt item\#1} (but not the color and the type).

The factored representation also introduces a factored learning problem: instead of learning a monolithic neural network for $\gT_\theta$ and $\textit{eval}_\theta$, the problem is factored into learning the grounding of individual predicates that appear in goal formulas, as well as the transition function for individual factors that were changed by an action.

\subsection{Representation Language}
\label{sec:pdsketch}
The overall specification of $\textit{eval}_\theta$ and $\gT_\theta$ can be decomposed into two parts: 1) the locality and sparsity structures and 2) the actual model parameters, $\theta$, such as neural network weights.  We provide a symbolic language for human programmers to specify the locality and sparsity structure of the domain and methods for representing and learning $\theta$.   If the human provides no structure, 
the model falls back to a plain object-centric dynamics model~\cite{zhu2018object}. However, we will show that explicit encoding of locality and sparsity structures can substantially improve the data efficiency of learning and the computational efficiency of planning with the resulting models. 

\model is an extension of the planning-domain definition language~\citep{fikes1971strips,fox2003pddl2}, a widely used formalism that focuses exposing locality and sparsity structure in symbolic planning domains. The key extensions are 1) allowing vector values in the computation graph and 2) enabling the programmer to use ``blanks'', which are unspecified functions that will be filled in with neural networks learned from data. Thus, rather than specifying the model in full detail, the programmer provides only a ``sketch''~\citep{solar2008program}.
The two key representational components of PDDL are predicates and action schemas (operators).

\begin{figure}[!tp]
    \centering
    \includegraphics[width=\textwidth]{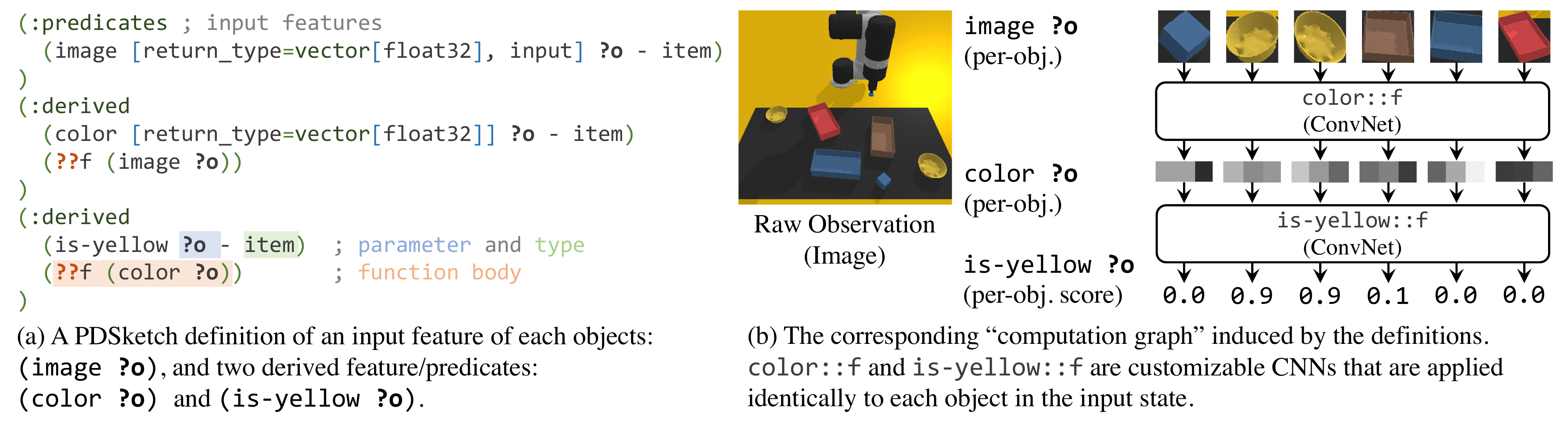}
    \vspace{-1.5em}
    \caption{A minimal example of defining derived features and predicates with blanks ``\slot''.}
    \label{fig:simplest-example}
\end{figure} 
\fig{fig:simplest-example} shows a simple example of PDSketch definition of predicates. All three predicates take a single object {\tt ?o} of type {\tt item} as their argument, and return either a floating-point vector or a scalar value from 0 to 1, indicating the score of a binary classifier. 
The {\tt image} predicate simply refers to the raw image crop feature of the object.  
The {\tt is-yellow} predicate's $\textit{grounding}$ takes a very simple form ``{\tt (\slot{f} (color ?o))}''. The term {\tt \slot{f}} defines a slot whose name is {\tt f}. It takes only one argument, the {\tt color} of the object {\tt ?o}, and outputs a classification score, which can be interpreted as the score of the object {\tt ?o} being yellow.  The actual computation of the {\tt yellow} predicate from the {\tt color} value (as well as the computation of the {\tt color} value from the {\tt image} value) is instantiated in a neural network with trained parameters. The computation graph for the whole model can be built by recursively chaining the function bodies of predicate definitions.

Next, we illustrate how {\it locality} and {\it sparsity} structures can be specified for an action schema. 
\fig{fig:structure} defines an action schema $\textit{name}$ {\tt move-into} with two effect components. First, highlighted in blue, the action changes the pose of object {\tt ?o1} to a new pose that depends on the current pose of the second object {\tt ?o2}. Rather than hand-coding this detailed dependence, we leave the grounding blank. In addition, in our domain, the {\tt wetness} of an object may be changed when the object is placed into a specific type of container. This is encoded by specifying a conditional effect using the keyword {\tt when}, with  two parts: 1) a Boolean-valued condition {\tt g} of some other predicates on the state (in this case, the types of the two objects), and 2) the actual ``effect'', in this case, to change the {\tt wetness} of {\tt ?o1} based on a function that considers the current {\tt wetness} of {\tt ?o1}. The update will be applied only if the condition is true. To ensure the computation is differentiable, we make this condition ``soft'': let $w$ be the current wetness, $w'$ be the new wetness computed by function {\tt ??h}, and $c$ be the scalar condition value computed by function {\tt ??g}. The updated value of the wetness will be $c w' + (1 - c) w$. Note that all the pose, wetness, and type representations can be arbitrary latent vectors computed by an encoder from the raw input. Thus, the ``blanks'' {\tt ??f}, {\tt ??g}, {\tt ??h} are indeed general neural networks. The effect definition here induces a corresponding computation graph of neural network weights and state representation tensors.

Like PDDL, PDSketch has full support of  first-order logic, including Boolean operations (and, or, not) and finite-domain quantifiers ($\forall$ and $\exists$). They allow us to define more complex structures in the domain of interest. We present our full language and more examples in the supplementary material.

\subsection{Model Learning and Planning with PDSketch}
Let $\theta$ denote the collection of all learnable parameters required to complete a  PDSketch domain definition into a full model.  This includes the parameters of the state encoder, all predicate groundings, and the definitions of slot-update functions that were left blank in the sketch.  
Recall that our training data are tuples of three sequences and a goal formula $\langle \overline{s}, \overline{a}, g, \overline{\textit{succ}}\rangle$.  Fundamentally, our objective is to minimize a sum of two losses, one related to predicting the truth values of the goal formulas and one related to predicting the next state, given the previous state and action:
\[\small \mathcal{L}(\theta) = \sum_{\langle \overline{s}, \overline{a}, g, \overline{\textit{succ}}\rangle \in \gD} \left\{\sum_{i} \text{BCE}\left( \textit{eval}_\theta(g, \gE_\theta(s_i)), \textit{succ}_i\right) + \sum_i \text{L1}\left( \gT_\theta(\gE_\theta(s_i), a_i), \gE_\theta(s_{i+1}) \right) \right\}, \]
where $\text{BCE}$ is the binary cross-entropy classification loss and $\text{L1}$ is a regression loss.  To avoid degenerate local optima, we add a ``lookahead'' loss term that combines both aspects,
as detailed in the supplement.  If the encoder $\gE_\theta$ is constrained to be the identity, and there is no predicate-level structure,  then the transition model essentially learns to be a per-object next-image predictor.  More generally, including the encoder turns this into a bisimulation objective~\citep{li2006towards}: we want to uncover a latent transition model that accurately predicts the reward signal (in this case, whether the goal is satisfied or not) but does not necessarily reconstruct the input state representation.

\begin{figure}[!tp]
    \centering
    \includegraphics[width=\textwidth]{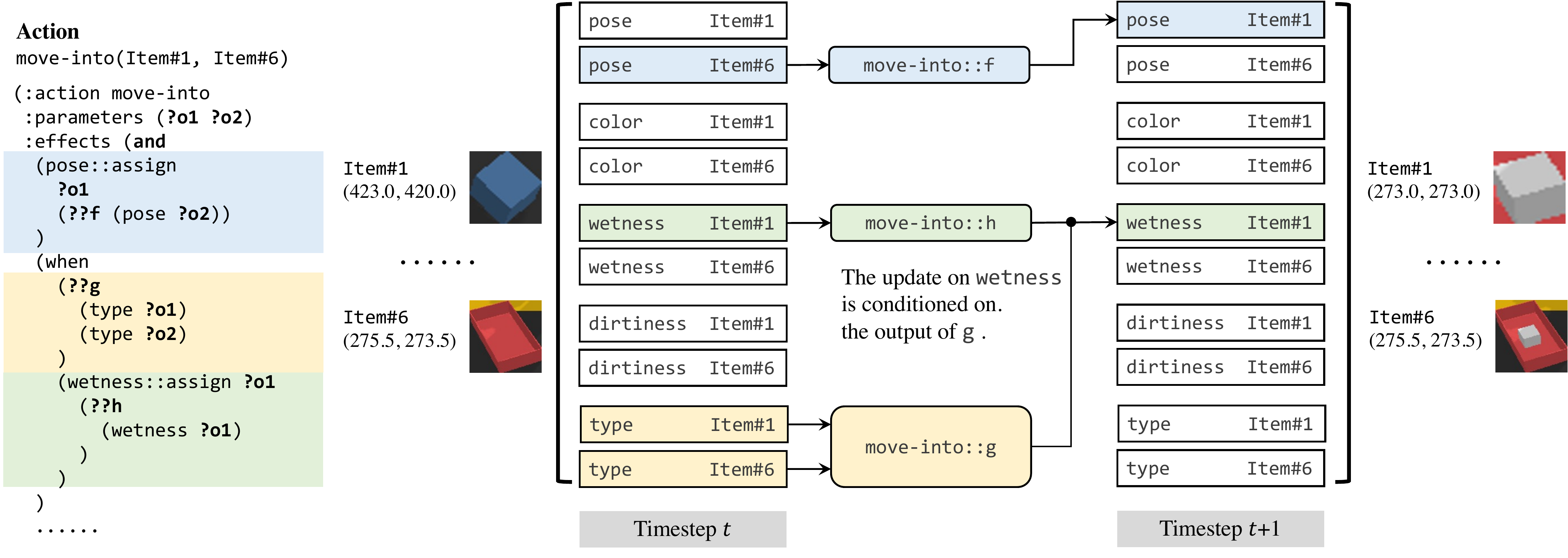}
    \vspace{-1em}
    \caption{A computation graph of a partial definition of the action ``move-into.''}
    \label{fig:structure}
\end{figure} 
In order to adjust $\theta$ to minimize this loss, we must establish a differentiable computation graph.  The encoder $\gE_\theta$ will generally be a relatively standard combination of convolutional and fully-connected feed-forward neural network.  
\fig{fig:structure} illustrates the computation graph associated with $\gT_\theta$ for a particular choice of action $a_i$ and objects that serve as its arguments. Importantly, note that there is a substantial amount of parameter-tying: the same predicate-grounding network might appear in multiple times, even when characterizing $\gT_\theta$ for a single action, if that predicate appears multiple times (applied to different objects) in the preconditions or effects of the action.  The computation graph shown here, as well as those necessary to compute $\textit{eval}_\theta(g, \gE_\theta(s_i))$, involve Boolean operators, which do not have useful derivatives for optimization. We address this by representing truth values as elements of the interval $[0, 1]$ and approximate logical operations with the differentiable Gödel t-norms: $\texttt{not}(p) = 1 - p$, $\texttt{and}(p_1, p_2) = \min(p_1, p_2)$, $\forall x.p(x) = \min_x p(x)$.

Once we have estimated $\theta$ from data, we can solve any planning problem in the domain given any starting state $s_0 \in \gS$ and goal $g$ expressed in terms of predicates for which we have groundings.  The resulting transition model $\gT$ can be used by a variety of different planners.  We will focus on forward search, constructing a tree rooted at $\gE(s_0)$ with branches corresponding to the possible instantiations, $a$,  of action templates $\gA$ with the objects in the universe $\gU$ associated with $s_0$, and next latent states computed by applying $\gT$.  The search terminates when it reaches a node $n$ in which $\text{eval}_\theta(n, g) > .5$.
Unguided forward search can be very slow when the planning horizon is long or the branching factor is large (\eg, when there are many objects in an environment). To address this, we will use the \astar heuristic search algorithm using a {\em domain-independent search heuristic} directly derivable from the locality and sparsity structure defined in PDSketch.

\subsection{Inducing Domain-Independent Heuristics}
Because of the rich representational capacity of PDSketch models, in which values can be continuous and multidimensional, we cannot take advantage of the 
planning algorithms that operate on PDDL input, such as Fast Downward~\citep{helmert2006fast}, which derive much of their efficiency from domain-independent heuristics.  We cannot
use their strategies in detail, but we take inspiration from the idea of deriving an optimistic estimate of the cost to reach the goal from a state by solving a ``relaxed'' version of the problem, which is computationally easier than the original~\cite{bonet2001planning}.

One way to construct a relaxed planning problem is to allow each predicate instance to take on multiple values at the same time. For example, the robot or an object can effectively be in multiple places at the same time.  
In this relaxation, computing the number of steps needed to change the value of a predicate instance can be done in polynomial time.   We can use such a relaxation in the hFF heuristic~\citep{hoffmann2001ff} to get an estimate of the cost to goal, by first chaining the actions forward until all the components of the goal condition have been made true, and then searching to recover a small set of actions that can collectively achieve the goal under the relaxation.  

To use hFF, however, we must reduce our continuous-space problem to a discrete-space one.
\newtext{Specifically, we discretizes all continuous state variables (poses, \etc) into a designated number of bins (\eg, 128). Next, for each externally-defined functions, we learn a first-order decision tree to approximate the computation (\eg, to approximate the neural network). Concretely, we use VQVAE~\citep{van2017neural} to discretize the feature vectors. For all predicates whose output is a latent embedding, we add a vector quantization layer after the encoding layers. We initialize the quantized embeddings by running a K-means clustering over the item feature embedding from a small dataset, and finetune the weights on the entire training dataset for one epoch.}
We describe this in more detail in the supplementary material. Although our discretizations are inherently lossy on non-Boolean predicate values, since they are only used for search guidance, and not in the forward simulation of the actual actions during planning, this does not affect the correctness of the overall algorithm.

\section{Experiments}
\label{sec:experiment}

\begin{table}[tp]
\vspace{0em} %
\begin{minipage}{0.27\textwidth}
    \centering\small
    \vspace{0em} %
    \includegraphics[width=0.8\textwidth]{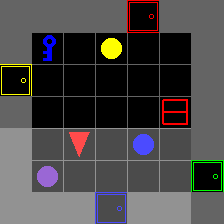}
    \vspace{0.2em}
    \captionof{figure}{A screenshot of the BabyAI environment.}
\end{minipage}\hfill
\begin{minipage}{0.7\textwidth}
    \centering\small
    \vspace{-2em}
    \setlength{\tabcolsep}{3pt}
    \begin{tabular}{l cccc cc}
    \toprule
        \multirow{2}{*}{Model} &  \multicolumn{4}{c}{Inductive Biases} & \multicolumn{2}{c}{Succ. Rate}\\ \cmidrule{2-5} \cmidrule{6-7}
        & \mycellc{Obj-Centric.} & Facing & \mycellc{Rob. Dyn.} & \mycellc{Obj. Prop.} & Basic &  \mycellc{\#Obj.Gen.} \\
    \midrule
        BC       & Y & N & N & N & 0.93 & 0.79 \\
        DT (S)      & Y & N & N & N & 0.91 & 0.82 \\
        DT (S+F) & Y & N & N & N & 0.32 & 0.19 \\ \midrule
        \newtext{DreamerV2} & N & N & N & N & \newtext{0.96} & \newtext{0.79} \\ \midrule
        PDS-Base & Y & N & N & N & 0.82 & 0.62 \\
        PDS-Abs  & Y & Y & N & N & 0.99 & 0.98 \\
        PDS-Rob  & Y & Y & Y & N & \textbf{1.00} & \textbf{1.00} \\
        \bottomrule
    \end{tabular}
    \vspace{0.2em}
    \caption{The planning success rate of different models on BabyAI.}
    \label{tab:babyai-main}
\end{minipage}
\vspace{-1.5em}
\end{table}

\begin{figure}[tp]
    \centering\small
    \includegraphics[width=\textwidth]{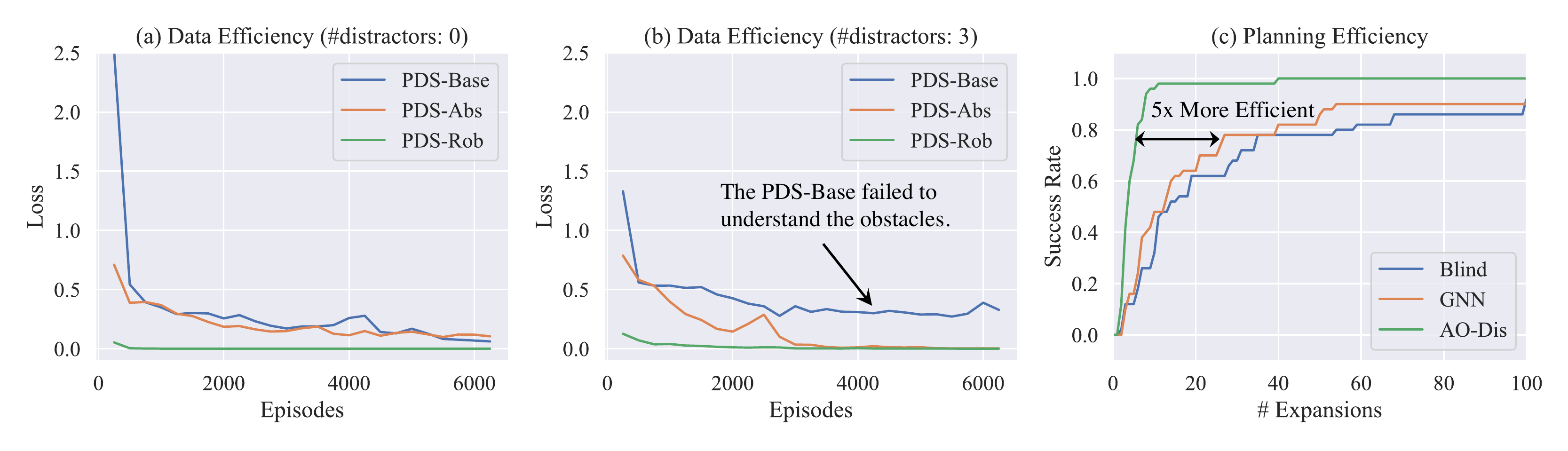}
    \vspace{-2.5em}
    \caption{(a) and (b) Data efficiency comparison for model learning. We compare three structures with different levels of abstractness. (c) Planning efficiency, measured as the number of expanded nodes for different heuristic computation methods.}
    \label{fig:babyai-figures}
\end{figure}
We evaluate PDSketch in two domains: BabyAI, a 2D grid-world and Painting Factory, a simulated tabletop manipulation task.
We compare our model with two model-free methods: Behavior Cloning~\citep[BC;][]{bain1995framework} and Decision Transformer~\citep[DT;][]{chen2021decision}. We implement two DT variants: DT-S with only successful demonstrations, and DT-S+F with successful and failed demonstrations. 
We use graph neural networks~\citep{gori2005new,battaglia2018relational} as their state encoder. 

\subsection{BabyAI}
BabyAI~\citep{babyai_iclr19} is an image-based 2D grid-world environment where an agent can navigate around obstacles, pick and place objects, and toggle doors. In this paper, we focus on a specific level of BabyAI, namely {\it ActionObjDoor}. At this level, the agent navigates within a 7x7 grid. The goals include {\it go to an X}, {\it pick up an X}, {\it open an X}, where {\it X} is a noun phrase, such as ``blue key''. We train all models on environments with 4 doors and 4 objects. The offline dataset $\gD$ contains both successful and failure demonstrations obtained by \astar search. We extend \model to interactive data gathering in the supplementary material. Additionally, we test generalization to environments with 6 doors and 8 objects. Since objects may block agents, the agent needs to successfully uncover the underlying dynamics and plan to navigate around them.

\begin{figure}[tp]
    \centering
    \includegraphics[width=\textwidth]{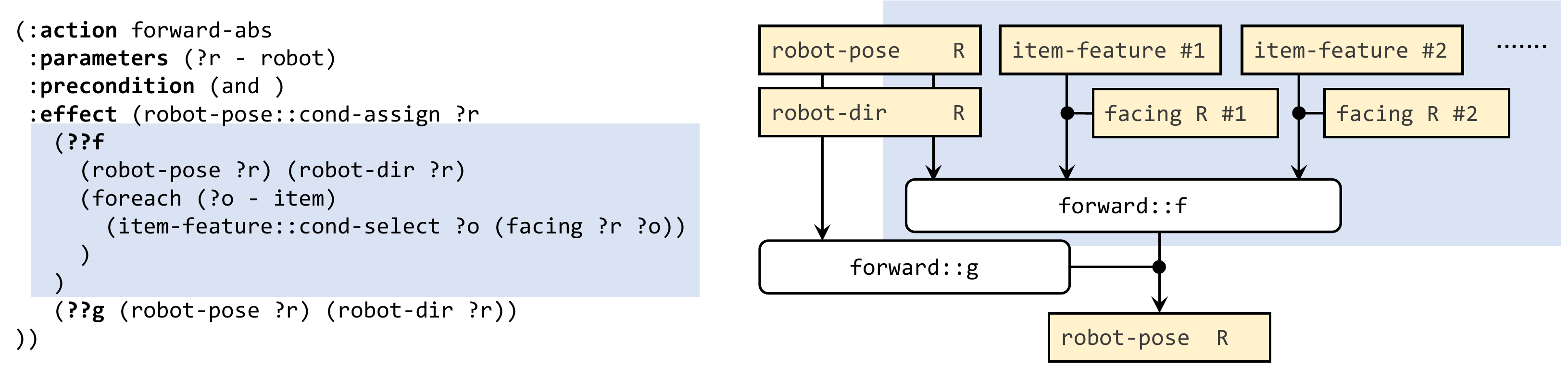}
    \caption{\newtext{The PDSketch definition of the ``forward'' action and the corresponding computation graph for the PDS-Abs model. The mentioned predicate {\tt facing} is an derived predicate represented as a neural network that is jointly learned (omitted in the graph for brevity).}}
    \label{fig:rob-abs}
\end{figure}

We study three \model models with various levels of {\it locality} and {\it sparsity} structures built in. The PDS-Base model has no built-in structures: each object is represented as a holistic vector. This falls back to an object-centric transition model~\citep{zhu2018object}. PDS-Abs model that disentangles poses from object appearances. Importantly, it defines a predicate {\it facing} and uses this concept to write action definitions (\eg, whether a robot move will be blocked by the object it is facing). However, the grounding of {\it facing} is to be learned. Figure~\ref{fig:rob-abs} shows the detail definition of the forward action in the PDS-Abs model. PDS-Rob contains predefined rules for robot movements but the object recognition modules are learned. We provide their definitions in the supplementary material.

\xhdr{Results.} \tbl{tab:babyai-main} shows the results. Overall, \model with more structure (PDS-Abs and PDS-Rob) outperforms baselines by a significant margin. From the performance of PDS-Abs, we see that even a tiny amount of additional structure (\eg, an {\it ungrounded} predicate ``facing'') significantly improves the performance, especially when it comes to generalization to more complex environments. See below for a zoomed-in analysis for model learning.
Furthermore, we hypothesize that the inferior performance of decision transformer in the mixed training data (Succ+Fail) setting is due to the reward sparsity: the agent only gets reward 1 when it reaches the goal. Thus, the failed demonstrations are generally hard to model as they are irrelevant to the goal specification. \newtext{In addition, we compare our models with DreamerV2, a state-of-the-art model-based reinforce learning algorithm for image-based environments. Compared with BC and DT, we see DreamerV2 achieves slightly improved performance for the basic task, but does not show stronger generalization to environments with more objects. We hypothesize this is because Dreamer still learns a fixed policy for execution.}

\xhdr{Data efficiency.} \fig{fig:babyai-figures}a-b shows model learning curves for the three \model models, in the cases where there is a single object in the environment and when there are 4 objects. . When the number of object is small, the environmental dynamics is easier to learn: both Base and Abs have a similar performance. However, when the number of object increases, \model can leverage the inductive bias to learn the dynamics faster. Shown in the figure, even at the end of training, the model PDS-Base has not successfully learned the correct movement dynamics, leading to its inferior performance during generalization to more objects.

\xhdr{Planning runtime efficiency.} \fig{fig:babyai-figures}c quantifies the number of nodes expanded by \astar when using different heuristic computations.
Specifically, we compare our model with the {\it blind} heuristic (\ie, the heuristic value of a state is 0 when it satisfies the goal and otherwise 1.) and a GNN-parameterized heuristic learned from successful demonstrations.
All results are based on the PDS-Abs model. First, learning-based heuristics (GNN) shows improvement over the baseline ``blind'' heuristic, especially when the task is easy (requiring a small number of expansions). Second, the heuristic derived from \model significantly improves the search efficiency. At the success rate of 0.8, our method is 5 times more efficient than learning-based heuristics.

\begin{figure}[tp!]
\begin{minipage}{0.6\textwidth}
\centering\small
    \includegraphics[width=\textwidth]{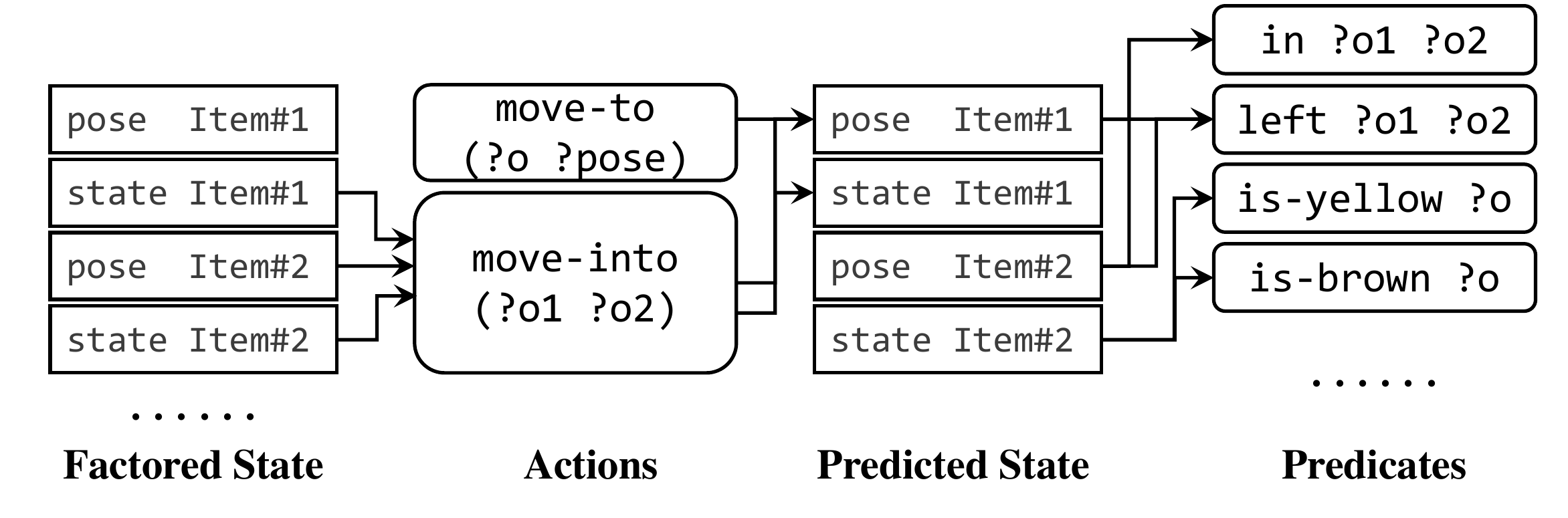}
    \vspace{-2em}
    \caption{Our encoding of the painting factory domain. Each object has a pose and a state. Action {\tt move-to} directly changes the object pose. {\tt move-into} moves an object into a container, which possible changes the state (\eg, color) of the object. Goals are represented using relational predicates.}
    \label{fig:painting-structure}
\end{minipage}
\hfill
\begin{minipage}{0.38\textwidth}
\centering\small
\setlength{\tabcolsep}{8pt}
\begin{tabular}{lcc}
\toprule
    Model & \mycellc{Succ. Rate\\@100} & \mycellc{Succ Rate\\@1000} \\ \midrule
    BC  & 0.70 & 0.95 \\
    DT  & 0.56 & 0.97 \\
    PDS & {\bf 0.91} & {\bf 0.99}\\
\bottomrule
\end{tabular}
\vspace{0.5em}
\captionof{table}{Model performance on Painting Factory, measured as plan success rate. The number after @ is the number of demonstration trajectories. \model models show stronger data efficiency than baselines.}
\label{tab:painting-result}
\end{minipage}
\vspace{1em}

    \centering\small
    \vspace{-0.5em}
    \includegraphics[width=\textwidth]{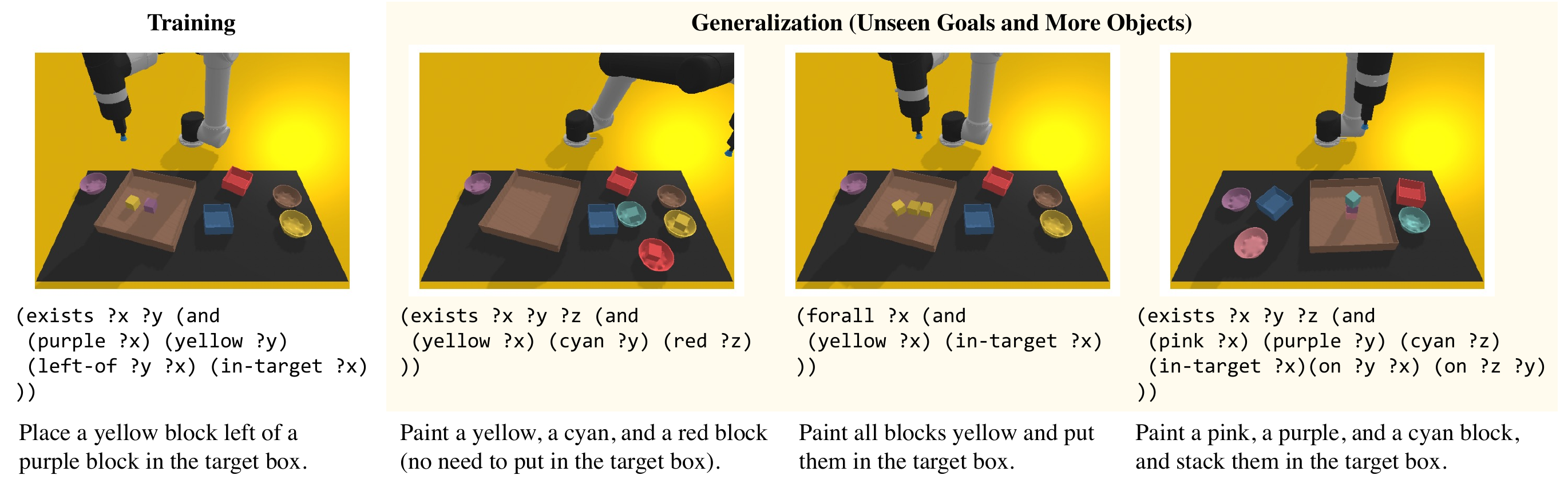}
    \vspace{-1em}
    \caption{From only one training task (left), a model specified in \model learns primitive concepts including object colors and spatial relations. They can be used in planning for unseen tasks given new first-order logic descriptions of the goal. The natural language descriptions are shown for readability.}
    \label{fig:painting-generalization}
\end{figure}

\subsection{Robot Painting}

Finally, we extend the framework to a tabletop manipulation task, \newtext{built based on the tabletop environment of \citet{zeng2020transporter}}. There are three zones, several bowls, and several blocks on the table. The robot can use its suction gripper to pick-and-place objects into a designated location. Blocks have 8 possible colors. Placing objects in a bowl will paint the object to be the same color as the bowl.
The task is to paint the blocks and organize them in the target brown box.
Our training-time goal requires the robot to paint-and-place two objects. The goal contains their colors and their relationship (\eg, a {\it pink} block is {\it left of} a {\it yellow} block. Demonstration are collected using hand-crafted oracle policies following~\citet{zeng2020transporter}. The offline dataset contains only successful trajectories. There are two built-in actions that the robot can execute.
\fig{fig:painting-structure} shows the computation graph derived from \model.

\xhdr{State and action space.} \newtext{The state space in the painting factory is composed of a list of objects, and a list of containers. Both objects and containers are represented as a tuple of a 3D xyz location, and a image crop. The image crop is generated by first computing the 2D bounding box of the object in the camera plain, cropping out the image patch, and resizing it to 32 by 32. The action space contains two primitives. First, {\tt move-into} is an action defined for each pair of item and container. It changes the pose of the item (now the item will be inside the container). When the item is placed into a bowl, it will be painted into the same color as the bowl. The second action takes an object and a 3D location as input, and moves the object directly to a designated position (ignoring the rotation).}

\xhdr{Results.}
\tbl{tab:painting-result} shows the planning success rate {\it on the training task} with different amounts of training data. Overall,  \model is more data efficient than both baselines and achieves strong overall performance in this task. Much more importantly, \fig{fig:painting-generalization} shows our {\it generalization to novel task specifications}, which involve more objects or new specifiers (\eg, {\tt forall}). Our model generalizes directly to these novel scenarios {\it without any additional training}. \newtext{The quantitative performance for these three generalization tasks are: 0.99, 0.98, and 0.87, respectively, measured as success rate after executing the plan. The last task has lower success rate because stacking objects may fail due to controller and physical noises. Future work may consider building closed-loop controller that can recover such failures.} We specify implementation details in the supplementary material.
\section{Related Work}
Integrated model learning and planning is a promising strategy for
building robots that can generalize to novel situations and novel goals. Specifically, \citet{chiappa2017recurrent,zhang2020learning,schrittwieser2020mastering} learn dynamics from raw pixels; \citet{jetchev2013learning,pasula2007learning,konidaris2018skills,chitnis2021learning,bonet2019learning,asai2020learning,silver2021learning} assume access to the underlying factored states of objects, such as object colors and other physical properties.
Our work bridges the gap between two groups: we do not assume pre-factored state representations given as the input, but learn to ground different factors of object states. More importantly, instead of relying on off-the-shelf models for predicting pixels or for learning first-order rules in symbolic domains, our work considers how human-programmed locality and sparsity structures can improve the model learning and planning efficiency.

Our model learns an object-factored transition model, which generally falls into the category of learning object-oriented MDPs~\citep[OO-MDPs;][]{guestrin2003generalizing,diuk2008object}. OO-MDPs have been applied to neural network-based representations for visual domains~\citep{walsh2010efficient,kansky2017schema,zhu2018object,xia2018learning,veerapaneni2020entity} and textual game domains~\citep{liu2021learning}. Our model 
leverages object-centric representations to encode permutation-invariant structures. Furthremore, we exploit fine-grained local and sparse structures in model learning.

The objective of model learning and planning is to obtain a goal-conditioned policy~\citep{kaelbling1993learning,kaelbling1993hierarchical}. While many others have studied model-free~\citep{dayan1992feudal,schaul2015universal} or hybrid model-free and model-based approaches~\citep{pong2018tdm,nasiriany2019planning}, in this paper, we focus on learning {\it structured} models from data and leveraging domain-independent planners for planning in latent representations. Such structured models support novel goal specifications via a first-order logic language, and domain-independent heuristics to accelerate search. \newtext{Another alternative approach towards learning and planning is to learning to predict subgoals that needs to be achieved before other goals~\citep{xu2019regression}. However, their work assumes that all preconditions and effects can be represented in a predefined symbolic language, and there are controllers for achieving individual subgoals. By contrast, \model supports generic neural-network-based representations for predicates and action effects.}

Our framework \model combines model definition in a structured language (\eg, first-order logic) and neural network learning. It is closely related to the idea of neuro-symbolic programming and differentiable programming for relational reasoning~\citep{manhaeve2018deepproblog,riegel2020logical,huang2021scallop} and policy learning~\citep{sun2019program}. Our language \model borrows ideas from earlier work on ``soft'' execution of logical formulas but works on planning domains.
\section{Conclusions and Limitations}
\label{sec:conclusion}
 \model supports flexible and effective specification of {\it locality} and {\it sparsity} structures of environment transition models. Leveraging these structures enables more data-efficient learning, compositional generalization to novel goal specifications and environmental states, and also domain-independent heuristics that accelerates performance-time planning.
  Limitations we hope to address include the lack of hierarchy and the inability of the method to discover novel factorizations.

\newtext{In terms of societal impact, PDSketch suggests a hybrid method for building intelligent robots, by combining
human programs to specify high-level structures with learning, to reduce the amount of data and computation required
for learning complex and long-horizon behaviors. It also enables more modularized systems: users
can specify new tasks using the predicates and actions that have already been defined more easily than in unstructured
approaches. However, PDSketch may require more expertise (or, at least, a different type of expertise) from programmers than other approaches.}

\vspace{1em}
\xhdr{Acknowledgement.}
We thank all group members of the MIT Learning \& Intelligent Systems Group for helpful comments on an early version of the project. This work is in part supported by ONR MURI N00014-16-1-2007, the Center for Brain, Minds, and Machines (CBMM, funded by NSF STC award CCF-1231216), NSF grant 2214177, AFOSR grant FA9550-22-1-0249, ONR grant N00014-18-1-2847, the MIT Quest for Intelligence, MIT–IBM Watson AI Lab. Any opinions, findings, and conclusions or recommendations expressed in this material are those of the authors and do not necessarily reflect the views of our sponsors.

{
\small
\bibliographystyle{plainnat}
\bibliography{reference}
}
\clearpage

\clearpage

\begin{center}
\bf \Large Supplementary Material for PDSketch:\\ Integrated Planning Domain Programming and Learning    
\end{center}

\appendix
The rest of the supplementary material is organized as the following. First, in \sect{secapp:results}, we provide additional results on data efficiency and on-policy learning for the BabyAI environment studied in the paper. Then, in \sect{secapp:language}, we present an example-based formal introduction of the \model language. Next, in \sect{secapp:heuristic}, we describe the implementation of our domain-independent heuristic. We also discuss why sparsity and locality structures are particularly important in inducing domain-independent heuristics. Finally, in \sect{secapp:setup}, we present our experiment setups. Both \sect{secapp:heuristic} and \sect{secapp:setup} are presented based on the language definition in \sect{secapp:language}.

\section{Additional BabyAI Results}
\label{secapp:results}
In this section, we supplement two additional results on the BabyAI environment. First, in \sect{secapp:dataeff}, we present the data efficiency study of different models in terms of their performance-time success rate. We also make variance analysis of different models across different runs. Next, in \sect{secapp:onpolicy}, we present how our model can be integrated with on-policy learning methods to make active data gathering.

\subsection{Data Efficiency and Variance Analysis}
\label{secapp:dataeff}
We further quantify the data efficiency of different models in terms of their performance-time success rate. Note that this is different from the Figure 7 in the main paper, where we have studied the data efficiency in terms of model learning accuracy. In this section, we show that, 1) compared to model-free baselines, integrate model learning and planning does improve the overall data efficiency in the domain we consider, and 2) the same is true for different \model models with different levels of abstractions and prior knowledge.

Concretely, we generate three datasets, with 100 demonstration episodes, 1000 episodes, and 10,000 episodes, respectively. We train the model on different datasets until converge, and evaluate their performance. We have repeated all experiments three times. The reported value are the average score and the standard deviation.

Results are summarized in \tbl{tab:data-efficiency-supp}. Here we summarize our findings as the following.

\begin{enumerate}
    \item First, in general, behavior cloning and decision transformer achieve similar performance on this task. When the amount of training data is small, the model perform noticeably worse than model-based methods.
    \item The base model, even if trained with a large amount of data, struggle to capture the robotic movement transitions when facing obstacles. This results in that the performance gets stuck at around 80\%.
    \item Compared to recognizing object properties, a significant amount of data is required to learn the transition model. When the transition model is given (in PDS-Rob), learning the visual recognition models in this simple grid-world domain is very data-efficient: using only 100 episodes, the model successfully solves 70\% of the tasks.
\end{enumerate}

\begin{table}[ht]
    \centering
    \small
    \setlength{\tabcolsep}{10pt}
    \begin{tabular}{lccc}
    \toprule
        Model & 100 episodes   & 1000 episodes & 10000 episodes \\\midrule
        BC       & 0.24 $\pm$ 0.01 & 0.24 $\pm$ 0.01 & 0.39 $\pm$ 0.03 \\
        DT       & 0.23 $\pm$ 0.02 & 0.24 $\pm$ 0.01 & 0.30 $\pm$ 0.05 \\
        PDS-Base & 0.04 $\pm$ 0.00 & 0.11 $\pm$ 0.01 & 0.80 $\pm$ 0.07 \\
        PDS-Abs  & 0.04 $\pm$ 0.02 & 0.98 $\pm$ 0.03 & \textbf{0.99} $\pm$ 0.01 \\
        PDS-Rob  & \textbf{0.72} $\pm$ 0.09 & \textbf{0.99} $\pm$ 0.01 & \textbf{0.99} $\pm$ 0.01 \\\bottomrule
    \end{tabular}
    \vspace{4pt}
    \caption{The planning success rate for different models using different amount of training data.}
    \label{tab:data-efficiency-supp}
\end{table}

\subsection{On-Policy Learning}
\label{secapp:onpolicy}
Our system can also be integrated into an on-policy learning setting. This is compatible with the standard model-based reinforcement learning setting. Specifically, recall that in an interactive learning setting, the system receives the current state and a goal expression. The agent itself should decide the next action to take. In the \model case, we run the planner based on the current model parameters $\theta$ to generate a plan. We follow the plan in the simulator and collect the resulting trajectory as well as the ``done'' signals from the environment. Finally, we update the model parameter $\theta$ using the newly collected data. Although our model may leverage offline data, to keep the algorithm simple, we do not implement data reusing. Within 10,000 episodes, our model PDS-Rob is able to reach 0.99 performance-time success rate. Comparatively, the reported performance of Proximal Policy Optimization (PPO)~\citep{schulman2017proximal,babyai_iclr19} needs 200,000 episodes.

We found that direct application of the on-policy algorithm on models with less inductive biases (\ie, PDS-Abs and PDS-Base) does not yield successful results. The success rate stuck around 6\%. This is primarily because of their failure in exploration: direct application of the planner generates successful trajectories at a very low probability. This can be potentially alleviated by adding random exploration factors or replanning when the agent fails to reach the goal.

Note that, since our model solely focuses on representing and learning the model, it is possible to integrate our framework with other model-based reinforcement learning algorithms, such as joint model and policy learning~\citep{schrittwieser2020mastering}, or world-model-based reinforcement learning~\citep{hafner2020mastering}. We leave these extensions as a future work.

\clearpage

\section{PDSketch Language}
\label{secapp:language}

In this section, we detail the design, the syntax, the semantics, and the implementation of the PDSketch definition language. We present it in two parts: the predicate and action definition (\sectapp{secapp:language-def}), and \model expressions (\sectapp{secapp:expression-def}).

\subsection{Predicate and Action Definition}
\label{secapp:language-def}
PDSketch is based on the Planning-Domain Definition Language~\citep{fikes1971strips,fox2003pddl2}, which is a language specialized and derived from LISP. We choose a LISP-style language for two important reasons. First, compared to other procedure- or object-centric languages such as Python, LISP features an easy definition of first-order logic, especially for recursively defined rules. Second, a significant amount of planning domains have been written in PDDL. Extending the language reduces the learning cost for programmers and also make old PDDL definitions easily reusable in PDSketch.

For readers who are familiar with PDDL, a PDSketch file is a domain definition file (in contrast to a problem definition file). PDSketch only contains the definition of predicates and actions in the domain. It does not contain problem-specific information, such as the current state of the world, or the goal specification.

Thus, a PDSketch file contains three parts. A PDSketch definition, at the highest level, takes the following form:

\begin{shaded}
\begin{verbatim}
definition:      "(" "define" "domain" domain-name-def content-def* ")"
domain-name-def: "(" "domain" string ")"
content-def:     type-def | predicate-def | derived-def | action-def
\end{verbatim}
\end{shaded}

An empty domain file can be written as:
\begin{shaded}
\begin{verbatim}
(define domain
  (domain my-domain-name)
)
\end{verbatim}
\end{shaded}
This defines a domain with name ``my-domain-name,'' with no predicates and actions defined.

A {\tt content-def} can be either a type definition, a predicate definition, or an action schema definition. First, type definition:
\begin{shaded}%
\begin{verbatim}
type-def:        "(" ":types" single-type-def ")"
single-type-def: type-name-list - base-type
type-name-list:  type-name+
type-name:       string
base-type:       type-name | prim-type | vector-type
prim-type:       "bool" | "int64" | "float32"
vector-type:     "vector" "[" prim-type (, int)? "]"
\end{verbatim}%
\end{shaded}

Below, we will be using the BabyAI world as an example, the example type definition of the BabyAI domain is:
\begin{shaded}
\begin{verbatim}
(:types
  robot item - object
  pose       - vector[float32, 2]
  direction  - vector[int64, 1]
)
\end{verbatim}
\end{shaded}

Specifically, a type definition contains multiple lines, each line being a pair of a type name list, and a base type. There are two types of types in PDSketch: object type and value type. Intuitively, object type refers to a concrete object in the world, while a value type usually denotes the return type of a feature function over the objects. Currently, we do not support the full hierarchical structure of type definitions: each type either inherits ``object'', which indicates this is an object type, or inherits a primitive value type (Boolean, integer, or floating-point numbers), or a vector type. In a vector-type definition, the second argument (\eg, the {\tt 2} in {\tt vector[float32, 2]}) denotes the dimension of the vector. This can be omitted.
Overall, the definition above defines four types: an object type called ``robot,'' an object type called ``item,'' a value type that is a 2D vector, named ``pose,'' and a value type that is a 1D vector of integer (\ie, a single integer), named ``direction.''

The {\tt predicate-def} and {\tt derived-def} are jointly used to define predicates. A predicate can be either a predicate directly observable from the environment, or a ``derived'' predicate, which is a predicate whose value is computed based on the value of other predicates. We first show the grammar definition.

\begin{shaded}%
\begin{verbatim}
predicate-def:        "(" ":predicates" single-predicate-def ")"
single-predicate-def: "(" predicate-name kwargs? variable-list ")"
predicate-name:       string
kwargs:               "[" kwarg "]"
kwarg:                kwarg-key "=" kwarg-value
kwarg-value:          "\"" string "\"" | int | bool | float | base-type
variable-list:        typed-variable*
typed-variable:       variable-name "-" type-name
variable-name:        string
\end{verbatim}%
\end{shaded}

\clearpage

An example definition of some input predicates is the following:
\begin{shaded}
\begin{verbatim}
(:predicates
  (robot-pose      [return_type=pose]            ?r - robot)
  (robot-direction [return_type=direction]       ?r - robot)
  (item-pose       [return_type=pose]            ?o - item)
  (item-image      [return_type=vector[float32]] ?o - item)
)    
\end{verbatim}
\end{shaded}
In this case, we have defined four predicates: a 2D position of the robot, the direction that the robot is facing, the 2D position of the item, and finally, an image representing a local crop of the item.

Based on the input predicates, users can define many ``derived'' predicates, whose values are automatically computed based on other input and derived predicates. The formal syntax is:
\begin{shaded}
\begin{verbatim}
derived-def: "(" ":derived" single-predicate-def expr ")"
\end{verbatim}
\end{shaded}
in which {\tt expr} is a syntax for expressions, which we will detail later. As an example, we look at the definition of several item property-related predicates.
\begin{shaded}
\begin{verbatim}
(:derived (item-feature [return_type=vector[float32, 64]] ?o - item)
  (??f (item-image ?o))
)
\end{verbatim}
\end{shaded}
This definition defines a predicate named {\tt item-feature}. It takes an item as its argument, and returns a 64-dimensional vector embedding associated with the object. Its expression is {\tt (??f (item-image ?o))}. That is, an unknown mapping from the input image of the item to a vector embedding. This definition will implicitly define a function whose name is {\tt derived::item-feature::f}. This function has no default implementation. The programmer, after loading the PDSketch definition, should register the corresponding implementation of this function. For example, one option is to associate this function with a convolutional neural network (CNN) that takes the image crop of each object {\tt ?o} and computes the corresponding item feature. Note that, in this definition, we are also implicitly defining a {\it parameter sharing} strategy: for all objects in the environment (independent of its absolute index), we will be applying the same CNN identically to them. We will detail possible expressions \model supports later. 

Based on this {\tt item-feature} definition, users can define several classifiers that will be useful in defining goal predicates.
\begin{shaded}
\begin{verbatim}
(:derived (is-red  ?o - item) (??f (item-feature ?o)))
(:derived (is-blue ?o - item) (??f (item-feature ?o)))
; ... more colors
(:derived (is-ball ?o - item) (??f (item-feature ?o)))
(:derived (is-door ?o - item) (??f (item-feature ?o)))
; ... more shapes
(:derived (is-open ?o - item) (??f (item-feature ?o)))  ; for doors
\end{verbatim}
\end{shaded}
By default, when the return type is not specified, the function returns Boolean values, which is consistent with the original PDDL syntax.

In BabyAI, items that the agent is holding has a special 2D position which is {\tt [-1, -1]}. Thus, whether the agent is holding an object can be classified by
\begin{shaded}
\begin{verbatim}
(:derived (robot-holding ?r - robot ?o - item)
  (??f (item-pose ?o))
)    
\end{verbatim}
\end{shaded}

Finally, we want to define a function indicating whether the robot is facing the object,
\begin{shaded}
\begin{verbatim}
(:derived (robot-facing ?r - robot ?o - item)
  (??f (robot-pose ?r) (item-pose ?o))
)
\end{verbatim}
\end{shaded}
In its expression, we are defining an unspecified function {\tt ??f}, which takes two arguments, the robot position, and the item position. With all these predicates, we can now define the tasks. Recall that the ActionObjDoor environment has the following three tasks:

\xhdr{go to a red box:}
\begin{shaded}
\begin{verbatim}
(exists (?o - item) (and
  (robot-facing agent ?o) (is-red ?o) (is-box ?o)
))
\end{verbatim}
\end{shaded}
\xhdr{pick up a red box:}
\begin{shaded}
\begin{verbatim}
(exists (?o - item) (and
  (robot-holding agent ?o) (is-red ?o) (is-box ?o)
))
\end{verbatim}
\end{shaded}
\xhdr{open a red door:}
\begin{shaded}
\begin{verbatim}
(exists (?o - item) (and
  (is-red ?o) (is-door ?o) (is-open ?o)
))
\end{verbatim}
\end{shaded}
In the first two goal specifications, we are using a name constant ``agent'' to denote the robot that we are considering.

Next, we are going to define actions. There are four actions the agent can perform. The syntax for action definition is:
\begin{shaded}
\begin{verbatim}
action-def: "(" ":action" (action-name kwargs)
  ":parameter" "(" variable-list ")"
  ":precondition" expr
  ":effect" expr
")"
\end{verbatim}
\end{shaded}
An action definition is composed of a name, a list of parameters, a precondition, and an effect. The parameter are the arguments to the action. The precondition is a Boolean output value, indicating the situation where this action can be executed. Note that, the precondition is not part of the transition model of the environment. Instead, it's a property that is associated with the underlying policy of this action. We will come back to this point when we discuss concrete examples. The effect is an expression that change the state variables according to certain rules.

Concretely, we start with a simple action: ``lturn,'' which models the behavior that the robot turns left.

\begin{shaded}
\begin{verbatim}
(:action lturn
 :parameters (?r - robot)
 :precondition (and )
 :effect (assign (robot-direction ?r) (??f (robot-direction ?r)))
 ; could also be written as:
 ; :effect (robot-direction::assign ?r (??f (robot-direction ?r)))
)
\end{verbatim}
\end{shaded}
This definition has defined an action named ``lturn,'' it takes a single parameter ``?r'' which indicates the robot that we want to control. It has an ``empty'' precondition, indicating that this primitive action can be executed anytime. Meanwhile, the effect is to change the ``robot-direction'' state variable associated with the robot {\tt ?r}. The rule is a based on an (unspecified) function that takes the current robot facing direction and returns the new direction: {\tt (??f (robot-direction ?r))}.

With first-order logic formula, one can define more complex actions. One good example is the definition of the ``forward'' action. In BabyAI, the forward action takes the following rule. When there is no object facing the agent, the agent can move forward. When there is an object facing the robot, the robot may be blocked by the object, in which case the robot position will not change. Note that not all map items will block robots' movement. Which type of objects will block the robot is to be learned by the learning algorithm. In this case, we can define the robot action in an intuitive way as the following:
\clearpage

\begin{shaded}
\begin{verbatim}
(:action forward
 :parameters (?r - robot)
 :precondition (and )
 :effect (robot-pose::assign ?r (??f
   (robot-pose ?r)
   (robot-direction ?r)
   (foreach (?o - item)          ; enumerate all items in the map
     (when (robot-facing ?r ?o)  ; if the robot is facing this item
       (item-feature ?o)         ; consider the item-feature of ?o
     )
   )
 ))
)
\end{verbatim}
\end{shaded}
We now break down this definition into pieces. First, this is an action that takes only one argument (the robot ``?r''). Similar to the case of ``lturn,'' this action does not have a precondition, indicating that the action can be applied in any situations. The effect definition for this action is slightly more complex. First, the action changes the state variable {\tt (robot-pose ?r)}. The new value is computed based on three things, the current {\tt robot-pose}, the current {\tt robot-direction}, and the feature of all objects {\tt ?o} that the robot is facing. Here, we are using two special keywords, {\tt foreach} and {\tt when}. The first one iterates over all items in the environment, and the second one selects the feature of object {\tt ?o} if the condition is satisfied. Importantly, in this case, the corresponding function {\tt f} should be implemented as a variable-length-input function. That is, depending on how many objects are selected, the number of input features will change. In our implementation, this function is implemented as a graph neural network (GNN). Of course, if the programmer knows more about the underlying transition model, they can specify more detailed structure, such as the following:
\begin{shaded}
\begin{verbatim}
; define a helper derived predicate.
(:derived (is-obstacle ?o - item) (??f (item-feature ?o)))
(:action forward-detail
 :parameters (?r - robot)
 :precondition (and )
 :effect (when
   ; when there is no item ?o such that
   ;   ?o is an obstacle and the robot is facing ?o
   (not (exists (?o - item) (and (is-obstacle ?o) (robot-facing ?r ?o)) ))
   ; the robot pose will move forward and the new pose
   ;   will be computed by the current pose and the facing direction.
   (assign (robot-pose ?r)
     (??f (robot-pose ?r) (robot-direction ?r))
   )
 )
)
\end{verbatim}
\end{shaded}

\clearpage

Similarly, we can have the following definition for ``the robot executes picking up.''
\begin{shaded}
\begin{verbatim}
; define a helper predicate indicating whether the object
;   can be picked up.
(:derived (can-pickup ?o) (??f (item-feature ?o)))
(:action pickup
 :parameters (?r - robot)
 :precondition (and )
 :effect (foreach (?o - item)
   (when (and (robot-facing ?r ?o) (can-pickup ?o) )
     (assign (item-pose ?o)
       ; a dummy function, that should be implemented to return 
       ;   [-1, -1], indicating the item is in robot's inventory.
       (??f )
     )
   )
 )
)
\end{verbatim}
\end{shaded}
Overall, the definition language allows users to encode various kinds of prior knowledge about the transition model into the representation. Meanwhile, it does not require human programmers to fully specify the details, especially the recognition and classification problems. The learning algorithm will follow the structure and learns the missing pieces.

Importantly, depend on the knowledge that the programmer has about the environment, such definition can be very detailed, or very abstract. At an extreme case, we can have the following definition:
\begin{shaded}
\begin{verbatim}
(:action mysterious-action
 :parameter (?r - robot)
 :precondition (and )
 :effect (and
   (robot-pose::assign ?r (??f1
     (robot-pose ?r) (robot-direction ?r) (item-pose ??) (item-feature ??)
   ))
   (robot-direction::assign ?r (??f2
     (robot-pose ?r) (robot-direction ?r) (item-pose ??) (item-feature ??)
   ))
   (foreach (?o - item) (item-pose::assign ?o (??f3
     (robot-pose ?r) (robot-direction ?r) (item-pose ??) (item-feature ??)
   )))
   (foreach (?o - item) (item-feature::assign ?o (??f4
     (robot-pose ?r) (robot-direction ?r) (item-pose ??) (item-feature ??)
   )))
 )
)
\end{verbatim}
\end{shaded}
Here we are using a syntax sugar: the notation {\tt (item-pose ??)} is equivalent to {\tt (foreach (?x - item) (item-pose ?x))}. That is, this function takes the pose of all items into consideration. Note that, in this case, we have {\it no} structure built in to the system: the action can change any state variable, and the change depends on all other variables in the state representation. In this case, depending on the actual implementations of {\tt f1} to {\tt f4}, this transition model can be implemented in any form, such as a graph neural network (thus an object-centric transition model).

\paragraph{Remark: Preconditions \vs conditional effects.} There are two seemingly similar ways to define the condition under which an action takes effect: preconditions and conditional effects. To better understand the different between these two options, consider the following two examples:
\begin{shaded}
\begin{verbatim}
(:predicates
  (p ?o - item)
  (q ?o - item)
)
(:action op1
 :parameters (?o - item)
 :precondition (p ?o)
 :effect (q:assign (??f (q ?o)))
)
(:action op2
 :parameters (?o - item)
 :precondition (and )
 :effect (when
   (p ?o)
   (q:assign (??f (q ?o)))
 )
)
\end{verbatim}
\end{shaded}
The key difference is that, a ``precondition'' is the property associated with the corresponding low-level controller for the action, that is $\pi_{\tt op1}$. Its semantics is that, the $\pi_{\tt op1}$ can only be executed when {\tt (p ?o)} returns true. It does not specify what will happen when we execute $\pi_{\tt op1}$ in such a situation. In contrast, the semantics of the second definition is that, the corresponding $\pi_{\tt op2}$ can be executed under any circumstances. However, only if the condition for the conditional effect is true (\ie, {\tt (p ?o)}, the state value for {\tt q} will be changed.

Thus, preconditions and conditional effects should be learned from different signals. Preconditions should be learned from the signal from the execution of the corresponding controller. More specifically, the controller $\pi_{\tt op1}$ should return a Boolean value indicating whether the controller can be executed. By contrast, the conditional effects should be learned from the outcome of the execution of the controller $\pi_{\tt op2}$.

Note that, the conditional effect formulation is the one that fits most of the concurrent environment setups for reinforcement learning. That is, the defined actions correspond to the primitive actions that the robot can execute. Thus, usually all actions are always applicable at any state (\ie, the precondition of all actions are empty).

\clearpage

\subsection{Expressions}
\label{secapp:expression-def}
\model has the following built-in operations.

\paragraph{Propositional logic operations: \texttt{and}, \texttt{or}, \texttt{not}, and \texttt{implies}.} The propositional logic operations take the following syntax:
\begin{shaded}
\begin{verbatim}
and-expr-def:     "(" "and"     expr* ")"
or-expr-def:      "(" "or"      expr* ")"
not-expr-def:     "(" "not"     expr  ")"
implies-expr-def: "(" "implies" expr expr ")"
\end{verbatim}
\end{shaded}
The semantics of these operators follows the convention of Boolean operations: conjunction, disjunction, negation, and implication. Their implementation are based on the differentiable Gödel t-norms: $\texttt{not}(p) = 1 - p$, $\texttt{and}(p_1, p_2) = \min(p_1, p_2)$, $\texttt{or}(p_1, p_2) = \max(p_1, p_2)$, and $\texttt{implies}(p_1, p_2) = \max(1 - p_1, p_2)$.

There are two conventional notations for Boolean operators used in the definition of action effects. First, when an action has multiple effects, they will be written within a big {\tt and} block. For example, the following definition:
\begin{shaded}
\begin{verbatim}
(:action set-pqr
 :parameters (?r - robot)
 :precondition (and )
 :effect (and
   (assign (p ?r) (??f))
   (assign (q ?r) (??f))
   (assign (r ?r) (??f))
 )
)
\end{verbatim}
\end{shaded}
Meanwhile, when a Boolean predicate is directly written in the effect, its semantics is that this Boolean state variable will be set to true. Similarly, when we write {\tt (not (t ?r))} where {\tt t} is a Boolean predicate, it is equivalent to {\tt (assign (t ?r) true)}.

\paragraph{Boolean quantification: \texttt{forall} and \texttt{exists}.} Similarly, the quantification operations are define as the following:
\begin{shaded}
\begin{verbatim}
forall-expr-def: "(" "forall" (typed-variable) expr ")"
exists-expr-def: "(" "exists" (typed-variable) expr ")"
\end{verbatim}
\end{shaded}
As an example, the following expression:
\begin{shaded}
\begin{verbatim}
(exists (?o - item) (is-red ?o))
\end{verbatim}
\end{shaded}
computes whether there is an item in the environment that is red.

Similarly, the implementation of Boolean quantifiers are: $\forall x.p(x) = \min_x p(x)$ and $\exists x.p(x) = \max_x p(x)$.

\paragraph{Assignment.} There is only one assignment operation
\begin{shaded}
\begin{verbatim}
assign-expr-def: "(" "assign" "(" predicate-name variable-list ")" expr ")"
\end{verbatim}
\end{shaded}
The first term is the target state variable, while the second term is the expression of the value. For example, {\tt (assign (p ?r) (??f))} means that the value of {\tt (p ?r)} will be assigned to the return value of function {\tt f}.

\paragraph{Foreach and condition.} There are two special operators: for-each and condition.
\begin{shaded}
\begin{verbatim}
foreach-expr-def: "(" "foreach" (typed-variable) expr ")"
when-expr-def:    "(" "when" expr expr ")"
\end{verbatim}
\end{shaded}

When they are used in a precondition or an expression (in contrast to being under the effect definition), they means:
\begin{itemize}
    \item {\tt foreach}: it selects all objects of the specified type. See the previous definition of action {\tt forward} as an example.
    \item {\tt when}: it specifies the scenario in which a state variable is relevant to the computation. See the previous definition of action {\tt forward} as an example.
\end{itemize}

When they are used in effect definitions, they means:
\begin{itemize}
    \item {\tt foreach}: it applies the effect formula to objects of the specified type. See the previous definition of action {\tt mysterious-action} as an example.
    \item {\tt when}: it indicates a conditional effect. The effect will be applied if and only if the condition is true. See the previous definition of action {\tt mysterious-action} as an example.
\end{itemize}

\paragraph{Syntax sugars.} \model includes the following syntax sugars:

~~~~{\tt (p::assign ?a1 ?a2 ... VALUE)} is equivalent to\\~~~~{\tt (assign (p ?a1 ?a2 ...) VALUE)}.\par
~~~~{\tt (p::cond-assign ?a1 ?a2 ... COND VALUE)} is equivalent to\\~~~~{\tt (when COND (assign (p ?a1 ?a2 ...) VALUE))}.\par
~~~~{\tt (p::cond-select ?a1 ?a2 ... COND)} is equivalent to\\~~~~{\tt (when COND (p ?a1 ?a2 ...))}.\par

\paragraph{Blank notation \slot.} The \slot operation can be used to replace any expressions. It takes the following syntax:
\begin{shaded}
\begin{verbatim}
slot-expr-def: "(" "??" predicate-name kwargs expr* ")"
\end{verbatim}
\end{shaded}
For example, the definition: {\tt (??f (item-feature ?o))} defines an function that will be externally implemented. It has name {\tt f}. It takes only one input, which is {\tt (item-feature ?o)}.

\clearpage

\section{Domain-Independent Heuristic}
\label{secapp:heuristic}
\begin{figure}[!tp]
    \centering
    \includegraphics[width=\textwidth]{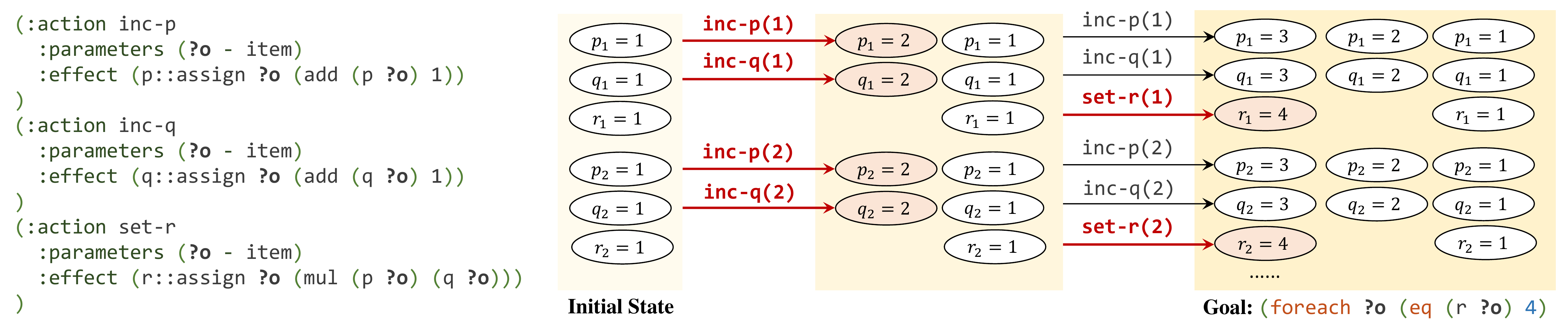}
    \caption{A graphical illustration of the hFF heuristics. The goal is to set the {\tt r} value of all objects to 4. The heuristic value $\text{hFF} = 6$, which is the number of selected operators (highlighted in red).}
    \label{fig:hff}
\end{figure}

In this paper, we focus on the hFF heuristic. Based on relaxed operators, it heuristically selects a set of operators that should be applied to accomplish the goal. Shown in \fig{fig:hff}, hFF sequentially tries to apply operators that yield novel values of the features. Once the goal condition is satisfied, it back-traces the used operators. Note that such computation is general and applies to domains with arbitrarily complex feature dependencies and number of objects. We present details of our implementation in the supplementary. However, such heuristics does not naturally work with vector embeddings and neural network predictors. Next, we talk about how to ``compile'' trained neural features and modules into hFF-compatible representations.

In this paper, we explore two strategies for performing this reduction. In the first, we simply change all the action models so that the result is to set all of the predicate instances that they effect to have a special value called {\em optimistic}, which is assumed to satisfy any goal test on that predicate instance.  (So, for example, after we have placed an object somewhere, we immediately "believe" it could be anywhere.)  An alternative strategy, which is more computationally complex but yields "tighter" (and therefore more helpful) heuristic values, is to explicitly discretize the value space of the predicates and directly reduce to the standard discrete case.  Both of these strategies are described in more detail in the supplementary material.  It is important to note that these methods are inherently lossy on non-Boolean predicate values, but since they are only used for search guidance, and not in the forward simulation of the actual actions during planning, they do not affect the correctness of the overall algorithm.

\paragraph{Optimistic compilation: leveraging locality.}
The first approach, optimistic compilation (OPT), compiles each non-Boolean state variable, \eg, {\tt (color ?o)}, into a boolean predicate {\tt color-opt}. Any operator that changes the value of {\tt color} (\eg, {\tt press-button}) will set {\tt color-opt} to true. Meanwhile, any Boolean expressions, such as {\tt (\slot~(color item\#1))} will return true as long as {\tt (color-opt item\#1)} is true. Intuitively, any operator that changes feature values will make the change {\it optimistically}.

The optimistic compilation ignores the fine-grained structure inside each state variables, but does leverage the object-level locality. For example, in the case of {\tt (forall ?o (eq (r ?o) 4)}, it will select two operators {\tt set-r(1)} and {\tt set-r(2)}, leading to $\text{hFF}=2$.

\paragraph{And-Or compilation.}
The And-Or compilation (AO) works in two steps. First, it discretizes all continuous state variables (poses, \etc) into a designated number of bins (\eg, 128). Next, for each externally-defined functions, it learns a first-order decision tree to approximate the computation (\eg, to approximate a neural network). Concretely, we use VQVAE~\citep{van2017neural} to discretize the feature vectors. For all predicates whose output is a latent embedding, \eg, the {\tt (item-feature ?o)} we introduced in our earlier example, we add a vector quantization layer after the encoding layers. We initialize the quantized embeddings by running a K-means clustering over the item feature embedding from a small dataset, and finetune the weights on the entire training dataset for one epoch.

For all predicates whose inputs and outputs are both continuous parameters. We use the FOIL algorithm, the first-order decision tree learning algorithm~\citep{quinlan1990learning} to extract first-order decision rules. We need to use FOIL instead of a plain (propositional) decision tree learning algorithm because the input to decision functions may have a variable number of objects. Here we are showing some examples of the quantized predicates:

\begin{shaded}
\begin{verbatim}
(is-red ?o - item) = (or
  (item-feature@3  ?o)
  (item-feature@14 ?o)
  (item-feature@16 ?o)
  (item-feature@10 ?o)
)
\end{verbatim}
\end{shaded}

Below we are showing a slightly more complex example, corresponding to the learned rule for action ``forward.''

\clearpage
\begin{shaded}
\begin{verbatim}
(((robot-pose ?r - robot) <- (SAS
  37 <- (or
  (and
    (not (robot-direction@2 ?r - robot))
    (not (robot-direction@3 ?r - robot))
    (not (robot-direction@1 ?r - robot))
    (robot-pose@37 ?r - robot)
  )
  (and (robot-direction@3 ?r - robot) (robot-pose@45 ?r - robot))
  (and (robot-pose@36 ?r - robot) (robot-direction@0 ?r - robot))
  (and (robot-pose@29 ?r - robot) (robot-direction@1 ?r - robot))
)
  4 <- (and (robot-pose@12 ?r - robot) (robot-direction@3 ?r - robot))
  40 <- (and (robot-pose@41 ?r - robot) (robot-direction@2 ?r - robot))
  21 <- (and (robot-pose@29 ?r - robot) (robot-direction@3 ?r - robot))
  32 <- (and (robot-pose@33 ?r - robot) (robot-direction@2 ?r - robot))
  3 <- (and
    (robot-pose@11 ?r - robot) (robot-direction@3 ?r - robot))
    (and
      (not (robot-pose@27 ?r - robot))
      (not (exists (_t0 - item) (and (robot-is-facing ?r - robot _t0 - item)
                                     (item-feature@9 _t0 - item))))
      (not (exists (_t0 - item) (and (robot-is-facing ?r - robot _t0 - item)
                                     (item-feature@0 _t0 - item))))
      (not (exists (_t0 - item) (and (robot-is-facing ?r - robot _t0 - item)
                                     (item-feature@6 _t0 - item))))
      (not (exists (_t0 - item) (and (robot-is-facing ?r - robot _t0 - item)
                                     (item-feature@11 _t0 - item))))
; ... more rules omitted.
\end{verbatim}
\end{shaded}
Note in the last section of the shown rules, the decision rule is written in first-order logic, where we need to quantify over objects.

The AO compilation further leverages the fine-grained structures of state variables. For example, the real value of features {\tt p}, {\tt q}, and {\tt r} will be discretized into bins, such as {\tt p=1}, {\tt p=2}, {\tt q=1}, {\tt q=2}. A decision tree will be used to represent the transition model. Thus, the resulting model will support fine-grained simulation and back-tracing.

\paragraph{The role of sparsity and locality structures in heuristics.} Although the discretization and heuristic computation itself does not assume any particular structure of the action definitions, their performance does rely on the programmed sparsity and locality structures, in particular, the ``factorization'' structure of the feature representation.

Specifically, without factorization, the entire state is described with one single vector embedding. This introduces a exponential number of possible states of the state. For example, let's assume an object state can be factorized into its x and y position (7 possible states per dimension), the color (6 possible states), and the shape (4 possible shapes). With factorization, we only need 24 possible state codes to represent each object. However, without factorization, we need $7 \times 7 \times 6 \times 4 = 1176$ states. Recall that, during the computation of heuristics, we will be solving a relaxed version of the planning program. If there is no factorization, the search problem reduces to a plain path-finding problem in the graph induced by the connectivity between these states, which is very inefficient to solve.

\clearpage

\section{Experimental Setup.}
\label{secapp:setup}
In this section, we detail the environment setup and domain definition of two environments: BabyAI and Painting Factory.

\subsection{BabyAI}
\paragraph{Setup.} Following the original BabyAI setup, we use a 7 by 7 grid as the physical world. The agent is initially positioned at the center, all other object locations are randomly selected. 
\paragraph{Offline data.} The data contains both successful demonstrations and unsuccessful demonstrations. For successful ones, we use the grid-world \astar search to generate the optimal trajectory from the original agent position to the target. For unsuccessful demonstrations, we first choose an incorrect object, approach the selected object, and then runs 5 steps of random walk.
\paragraph{Baselines.} For all baselines and our models, we use the following ways to encode object states, robot positions, and facing directions.

\begin{enumerate}
    \item object state: following the FiLM model presented in \citet{babyai_iclr19}, we use an integer embedding for object states.
    \item object position and robot position: we directly use the raw value as the input to the neural networks. We have tried to use embedding based methods to encode the values, but have seen consistently worse performance.
    \item robot facing direction: there are four directions. We use a 4 learnable embeddings for them.
\end{enumerate}

Thus, the input to the baselines are: 1) the robot state embedding, 2) the task goal specification, encoded as the concatenation of three word embeddings: the verb, the adjective, and the noun, and 3) the object state embeddings.

For both baselines BC and DT, we use a two-layer graph neural network to encode the world state into a 128-dimensional vector embedding. For BC, we use a single linear layer to predict the action to take. For DT, we use a transformer layer to aggregate the history (especially encoding the cost-to-go), and use another linear layer to predict the action in the next take.

\paragraph{Models.} For models written in \model, we have three variants.

\begin{enumerate}
    \item In PDS-Base, the robot state encodings are treated as a single vector, named {\tt robot-feature}, and the item state encodings are also treated as a single vector (which is the concatenation of their positions and visual features). The action definition encodes no structures. For example,
\begin{adjustwidth}{-3em}{0em}
\begin{shaded}
\begin{verbatim}
(:action forward
 :parameters (?r - robot)
 :precondition (and )
 :effect (and
   (robot-feature::assign ?r (??f (robot-feature ?r) (item-feature ??)))
 )
)
\end{verbatim}
\end{shaded}
\end{adjustwidth}
\clearpage
    \item In PDS-Abs, we have the abstraction of the idea ``robot-facing.'' This predicate is used to define actions. For example,
\begin{adjustwidth}{-3em}{0em}
\begin{shaded}
\begin{verbatim}
(:action forward
 :parameters (?r - robot)
 :precondition (and )
 :effect (robot-pose::cond-assign ?r
 (??f
   (robot-pose ?r) (robot-direction-feature ?r)
   (foreach (?o - item)
     (item-feature::cond-select ?o (robot-is-facing ?r ?o))
   )
 )
 (??g (robot-pose ?r) (robot-direction-feature ?r))
 )
)
\end{verbatim}
\end{shaded}
\end{adjustwidth}
    \item In PDS-Rob, the robot's movements have been encoded into the definition. For example, 
\begin{adjustwidth}{-3em}{0em}
\begin{shaded}
\begin{verbatim}
; The first function returns the 2D position that the robot is facing
;   when the robot is at ?p and facing ?d.
;   This function has been implemented externally.
; The second function returns whether an object is an obstacle (so
;   it will block the robot's movement.
;   This information has also been given to the agent as a input.
(:predicates
  (facing [return_type=pose] ?p - pose ?d - direction)
  (is-obstacle ?o - item)
)
(:derived (_is-facing ?p - pose ?d - direction ?t - pose)
  (equal (facing ?p ?d) ?t)
)
(:derived (_robot-facing [return_type=pose] ?r - robot)
  (facing (robot-pose ?r) (robot-direction ?r))
)
(:derived (_robot-is-facing ?r - robot ?o - item)
  (_is-facing (robot-pose ?r) (robot-direction ?r) (item-pose ?o))
)
(:derived (_robot-facing-clear ?r - robot)
  (not (exists (?o - item) (and 
    (_robot-is-facing ?r ?o)
    (is-obstacle ?o)
  ))
)
(:action forward
 :parameters (?r - robot)
 :precondition (_robot-facing-clear ?r)
 :effect (and (robot-pose::assign ?r (_robot-facing ?r)) )
)
\end{verbatim}
\end{shaded}
\end{adjustwidth}
\end{enumerate}

\subsection{Painting Factory}
\paragraph{State space.} The state space in the painting factory is composed of a list of objects, and a list of containers. Both objects and containers are represented as a tuple of a 3D xyz location, and a image crop. The image crop is generated by first computing the 2D bounding box of the object in the camera plain, cropping out the image patch, and resizing it to 32 by 32. The image encoder is a 3-layer convolutional neural network followed by a linear transformation layer into a 128-dimensional embedding. The xyz location are represented as three numbers, normalized into the range $[0, 1]$.

\paragraph{Action space.} The action space in the painting factory contain two actions. We first show the \model definition of two actions:
\begin{shaded}
\begin{verbatim}
(:action move-into
 :parameters (?o - item ?c - container)
 :precondition (and )
 :effect (and
   (item-pose::assign ?o (container-pose ?c))
   (item-feature::cond-assign ?o
     (??g (container-feature ?c))  ; when the container is a bowl,
     (??h (container-feature ?c))  ; paint ?o to be the same color as ?c.
   )
 )
)
(:action move-to
 :parameters (?o - item ?p - pose)
 :precondition (and )
 :effect (and (item-pose::assign ?o ?p))
)
\end{verbatim}
\end{shaded}
The first action, {\tt move-into} is an action defined for each pair of item and container. It changes the pose of the item (now the item will be inside the container). When the item is placed into a bowl, it will be painted into the same color as the bowl.
The second action moves the object directly to a designated position.

\paragraph{Offline data.} Recall that our task is to place two painted objects in the target location to form a designated relationship. To generate demonstrations, we first randomly select two blocks. Next, we paint one of them and place it at the center of the target location. Then, we paint the second object to the designated color and place it with respect to the first object.

\paragraph{Baselines.} For all baselines, we use a two-layer graph neural network (GNN) to encode the objects. We generate the action in two steps. First, for each (block, container) pair, we generate a scalar logit indicating the score of placing the block into the container. For each block, we also generate a scalar logit indicating the score of directly moving the block. We normalize these logits with a softmax to formulate the action probability. If the chosen action is to directly move the block, we also predict the target location.

\paragraph{Models.} The key feature in the painting factory domain is that now we need to plan for continuous actions, instead of a discrete space. This is handled using a basic task and motion planning strategy introduced in PDDLStream~\citep{garrett2020pddlstream}. We do this in two steps.

First, during training, we build a structured model for classifying geometric relationships (on, left, and right) in the following form:
\[ p(a~\text{\tt on }b) = \sigma\left( \frac{\|\textit{pose}(a) - \textit{pose}(b) - \theta\|^2_2 - \gamma}{\tau} \right), \]
where $\theta$, $\gamma$, $\tau$ are learnable parameters. $\sigma$ is the Sigmoid function. That is, intuitively, we say the block $a$ is on block $b$ if the pose of $a$ is close to $\textit{pose}(b) + \theta$. This formulation allows us to ``sample'' poses of block $a$ given the pose of block $b$.

Thus, during performance time, based on the initial pose of all objects, we sample a set of new poses that are on, left to, and right to the initial poses. We then use these poses to ground concrete operators. This strategy is also called ``incremental'' sampling in sampling-based task and motion planning literature.
 
\end{document}